\documentclass[sigconf]{acmart}
\usepackage{booktabs} % For formal tables
\usepackage{amsmath}
\usepackage{algorithm}
\usepackage[noend]{algpseudocode}
\usepackage{xspace}
\usepackage{enumitem}
\usepackage{color, colortbl}
\usepackage{hhline}

\definecolor{Gray}{gray}{0.8}

\def\Unanonymous{1}

\ifx\Unanonymous\undefined

\else

\fi

\begin{document}
\title {Personal Dynamic Cost-Aware Sensing for Latent Context Detection}
%\title{Addressing Cold Start in Consumption Intent Prediction for Sessions using Deep Learning}
%\titlenote{Produces the permission block, and   copyright information}
%\subtitle{Extended Abstract}
%\subtitlenote{The full version of the author's guide is available as
%  \texttt{acmart.pdf} document}

\ifx\Unanonymous\undefined

\author{Anonymous}
%\authornote{Dr.~Trovato insisted his name be first.}
\orcid{}
\affiliation{%
	\institution{Anonymous}
	%\streetaddress{P.O. Box 1212}
	\city{}
	\state{}
	\postcode{}
}
\email{}

\else

\author{Saar Tal}
%\authornote{Dr.~Trovato insisted his name be first.}
\orcid{1234-5678-9012}
\affiliation{%
	\institution{Ben-Gurion University of	the Negev}
	%\streetaddress{P.O. Box 1212}
	\city{Beer Sheva}
	\state{Israel}
	\postcode{43017-6221}
}
\email{asnatm@post.bgu.ac.il}

\author{Bracha Shapira}
%\authornote{Dr.~Trovato insisted his name be first.}
\orcid{1234-5678-9012}
\affiliation{%
	\institution{Ben-Gurion University of	the Negev}
	%\streetaddress{P.O. Box 1212}
	\city{Beer Sheva}
	\state{Israel}
	\postcode{43017-6221}
}
\email{bshapira@post.bgu.ac.il}

\author{Lior Rokach}
%\authornote{Dr.~Trovato insisted his name be first.}
\orcid{1234-5678-9012}
\affiliation{%
	\institution{Ben-Gurion University of	the Negev}
	%\streetaddress{P.O. Box 1212}
	\city{Beer Sheva}
	\state{Israel}
	\postcode{43017-6221}
}
\email{liorrk@bgu.ac.il}

\fi
\iffalse
\author{G.K.M. Tobin}
\authornote{The secretary disavows any knowledge of this author's actions.}
\affiliation{%
  \institution{Institute for Clarity in Documentation}
  \streetaddress{P.O. Box 1212}
  \city{Dublin}
  \state{Ohio}
  \postcode{43017-6221}
}
\email{webmaster@marysville-ohio.com}

\author{Lars Th{\o}rv{\"a}ld}
\authornote{This author is the
  one who did all the really hard work.}
\affiliation{%
  \institution{The Th{\o}rv{\"a}ld Group}
  \streetaddress{1 Th{\o}rv{\"a}ld Circle}
  \city{Hekla}
  \country{Iceland}}
\email{larst@affiliation.org}

\fi
% The default list of authors is too long for headers.
\renewcommand{\shortauthors}{Shapira et al.}

\begin{abstract}
In the past decade, the usage of mobile devices has gone far beyond simple activities like calling and texting. Today, smartphones contain multiple embedded sensors and are able to collect useful sensing data about the user and infer the user's context. The more frequent the sensing, the more accurate the context. However, continuous sensing results in huge energy consumption, decreasing the battery's lifetime. We propose a novel approach for cost-aware sensing when performing continuous latent context detection. The suggested method dynamically determines user's sensors sampling policy based on three factors: (1) User's last known context; (2) Predicted information loss using KL-Divergence; and (3) Sensors' sampling costs. The objective function aims at minimizing both sampling cost and information loss. The method is based on various machine learning techniques including autoencoder neural networks for latent context detection, linear regression for information loss prediction, and convex optimization for determining the optimal sampling policy. To evaluate the suggested method, we performed a series of tests on real world data recorded at a high frequency rate; the data was collected from six mobile phone sensors of twenty users over the course of a week. Results show that by applying a dynamic sampling policy, our method naturally balances information loss and energy consumption and outperforms the static approach.% We compared the performance of our method with another state of the art dynamic sampling method and demonstrate its consistent superiority in various measures. %Our methods outperformed , and were able to improve we achieved better results in either sampling cost or information loss, and in some cases we improved both. 

% Please include a maximum of seven keywords
%\keywords{keyword 1, \emph{keyword 2}, keyword 3, keyword 4, keyword 5, keyword 6, keyword 7}
\end{abstract}

\ccsdesc[500]{Information systems~Recommender systems}
\ccsdesc[500]{Computing methodologies~Neural networks}
\ccsdesc[300]{Computing methodologies~Supervised learning by classification}

\keywords{Context-aware recommendation algorithms, Mobile interface,  Cost-Aware Sensing}

\maketitle

\section{Introduction}
In the past decade, the area of context and context-aware computing has become the focus of much research \cite{Chen00,Perera14,BALDAUF07}. Currently, the use of mobile devices has gone far beyond simple activities like calling and texting. Today, smartphones contain multiple embedded sensors such as GPS, accelerometer, microphone, etc. that enable collection of data about the user and user's context inference \cite{sarker16}. Hence, context-aware systems can use this context to adapt their operations to the user without user intervention \cite{BALDAUF07}. Context-aware applications have been proposed in several domains including recommendation \cite{Setten04,Linas11,Alhamid16}, health-care \cite{Bardram04,Bardram03,Mitchell00,Kang06,Solanas14,Munuz03,Kjeldskov04}, smart homes \cite{kabir15,Skubic09}, data security \cite{Chakraborty13,Mirsky17,Muhtadi03}, etc. 

In the literature we can find many different definitions and perceptions regarding the term context. A user's context can be defined by his/her location, time of day, season, temperature, activity, environment, and even his/her emotions and mental state \cite{ABOWD99}.

Context inference can be divided into two categories: explicit or latent inference. Explicit context describes known user situations from a predefined set of contexts (e.g., "at work," "running") and hence can be better explained. However, it is challenging and a resource demanding task to define and train a large enough set of explicit contexts to cover the potentially large variety of user behaviors. Latent contexts are comprised of an unlimited number of hidden context patterns which are modeled as numeric vectors. They can be obtained automatically by applying unsupervised learning techniques on available raw data (e.g., mobile sensors)\cite{unger16}.

Context detection in mobile devices is done by analyzing data collected from the device's sensors. The more frequent the sampling (also referred to as sensing), the more accurate the context \cite{yuror13}. However, continuous sensing results in huge energy consumption, decreasing the battery's lifetime. Therefore, one of the main challenges in mobile context-aware applications is cost-aware accurate context detection where cost refers to the power consumption of the device \cite{sarker16}. When facing this challenge, a trade-off between the sensing accuracy and energy efficiency is required. 

We propose a novel approach for cost-aware sensing while performing continuous latent context detection. The suggested approach dynamically determines the user's sensors sampling policy (i.e. number of time intervals between samples for each sensor) based on three main factors: (1) the user's last known context - a latent context vector which is a reduced dimensional representation of user's features; (2) a supervised machine learning model which predicts information loss based on KL-Divergence between the actual context and the estimated context of candidate policies; and (3) sensors' sampling costs in terms of energy consumption. The main objective of the approach is to dynamically manage the trade-off between energy consumption and information loss, while minimizing both. The method is based on various machine learning techniques including autoencoder neural networks for latent context detection, linear regression for information loss prediction, and convex optimization for determining the optimal sampling policy. The objective function is nonlinear and takes into consideration both weighted predicted information loss and sampling costs. The models we use are user-personalized, trained and created for each user separately. In their paper, Lockhart et al.\cite{lockhart14} compared different learning algorithms for activity recognition on impersonal, personal and hybrid models. Their results show that the personal models outperforms the impersonal models.

While multiple studies have been conducted that suggested various approaches to deal with the energy-accuracy trade-off \cite{paek10,yuror11,yuror13,sarker16,zhang13,ben09,rachuri10a,rachuri10b,rachuri11,Nath12,kang08,Wang09}, none of them provide a solution that is designated for latent contexts. Furthermore, none of those methods exploit latent context and information loss when applying dynamic sampling for an unlimited number of sensors. Moreover, we are the first to present KL-Divergence as a measure for information loss in the task of context detection. 

The main contributions of this paper are as follows:
\begin{itemize}[noitemsep,nolistsep]
\setlength{\itemindent}{0em}
\item We suggest a novel energy efficient framework that utilizes dynamic sensing while taking both the user's last known latent context and predicted information loss into account.
\item The method we propose is generic and applicable for an unlimited number or type of sensors.
\item The proposed method is designated for latent context and is not limited to a finite set of contexts.
\item Since context is latent and comes in the shape of a numeric vector, we suggest KL-Divergence as a new measure for information loss in the task of latent context detection.
\end{itemize}

\section{Related Work}

Many studies have been conducted in the area of energy efficient context sensing. SenseLess \cite{ben09}, SensTrack \cite{zhang13} and RAPS \cite{paek10} are location detection applications that reduce the use of expensive sensors by using less expensive sensors more frequently. SenseLess \cite{ben09} uses the accelerometer sensor to trigger the more expensive GPS sampling when motion is detected. SensTrack \cite{zhang13} selectively executes GPS sampling based on acceleration and orientation sensors' data. In addition, when the user moves indoors and GPS is unavailable, it switches to Wi-Fi sensing method. RAPS \cite{paek10} uses the accelerometer to avoid GPS sampling when the user is stationary. Moreover, it uses the location-time history of the user to estimate the user's velocity and adaptively turns on the GPS only if the estimated uncertainty about the position exceeds the accuracy threshold. It also avoids turning on the GPS when it is not available and utilizes Bluetooth communication to reduce position uncertainty among neighboring devices. These methods provide a fine solution for the energy-accuracy trade-off, however their approach is limited to location detection and focuses on specific motion and location sensors without using machine learning techniques. In contrast, our approach is applicable for any sensors or latent context and automatically determines the sampling policy using machine learning techniques.

%Senergy\cite{Kansal13} is an API for developers that enables developers to control the latency (i.e., time between samples), accuracy, and battery (LAB) trade-off. The developer must deliver a set of explicit contexts to detect context (location, stationary, walking, and driving), while priority and quantity constraints on LAB factors are optional. The authors predefine the trade-offs based on experiments with different sets of sensors and contexts. Based on those trade-offs and the developer's specifications, the best set of sensors is selected. This method clearly considers the accuracy-energy trade-off. However, while their approach only applies on a small group of predefined contexts, our approach utilizes latent context. Furthermore, while their approach predefines the trade-offs and doesn't consider the difference in activities among different users, our method uses personalized models using machine learning to learn the trade-off.

Yurur et al. \cite{yuror11,yuror13} and Sarker et al. \cite{sarker16} adjust sampling frequency and duty cycle by measuring the stability of the sensors' values. Sampling frequency refers to the number of samples within a cycle, and duty cycle refers to the portion of time of an operational cycle spends on sampling. Duty cycle and sampling frequencies are chosen from a Cartesian product of two sets. While in their approach the sensing options are discrete, our method utilizes a continuous optimization function. Furthermore, while their technique cannot be adapted to sensors that don't support setting different sampling periods (GPS, Wi-Fi, etc.), our method is applicable for all sensors. Moreover, while their methods only take into account the sensor stability and don't consider the context, our method utilizes the context itself, as well as the predicted information loss which is more relevant to context-aware applications. 

Another approach for handling the accuracy-energy trade-off is presented by Rachuri et al. \cite{rachuri10a,rachuri10b}. They use adaptive sensor sampling which relies on the dynamic selection of predefined back-off/advance functions based on event history. The sampling interval decreases/increases when an interesting/non-interesting event is observed. Furthermore, depending on event stability, the method switches from the least to most "aggressive" function. While this approach sets the sampling interval at any interval the predefined functions supply, our approach determines sampling intervals which are the products of a minimal applicable sampling interval. Moreover, while the authors use the number of consecutive samples of the same state (interesting or not) to switch between functions in a conditional form, our approach considers the context itself and uses machine learning to learn the predicted information loss when using different policies.

In SociableSense \cite{rachuri11}, Rachuri suggests a different approach that adjusts the sensor's duty cycle according to its sensing probability. The sensing probability is dynamically calculated and defined as the portion of successes of previous sensing actions. A success indicates that the sensing action resulted in capturing an interesting event (domain dependent). The sensors are sampled at a high rate when there are interesting events observed and at a low rate when there are no events of interest. While their approach utilizes dynamic duty cycle for each sensor separately, our approach takes into consideration the combination of all sensors together when determining the sampling policy.

\section{Method}

We propose a novel approach for cost-aware sensing when performing continuous latent context detection. The proposed method is based on dynamic sampling policies. The sampling policy, which specifies when to resample each sensor, is determined dynamically according to the last known latent context, and the objective function aims at minimizing both sampling cost and information loss. Information loss is measured as the KL-Divergence between the actual user's context and his/her last known context, which are detected based on the actual and last known sensors' values respectively. The actual ground truth was created by frequently recording all sensors. We learned the difference in information loss between contexts that are inferred from the ground truth sensing and simulated scenarios. To create a simulated dataset that reflects the no-sensing scenario for different time periods, we added synthetic records with data from previously sensed records to the user's records. Thus, we were able to investigate the trade-off between energy efficient sensing and accurate context detection. 

When new data is received, we predict the information loss based on the learned differences and last known context. 

Our method consists of the following four steps (that will be explained in greater detail in the following sections): 
\begin{enumerate}
\item Data Extension: Multiple synthetic records are created from each existing record, such that sensors' values are taken from older records. Each record (synthetic or actual) and its corresponding distances between the actual record to the older records are maintained in the dataset.
\item Latent Context Detection: An autoencoder is trained to detect latent context on the raw sensor data by reducing the dimensionality of the sensor data and representing the latent context as a compact vector. When training is complete, latent context is detected for each record (synthetic or actual) in the extended dataset.
\item Information Loss Prediction: Information loss is calculated between each pair consisting of the synthetic record's context and the corresponding actual record's context. Afterwards, based on actual context, distances between samples, and corresponding information loss, a linear regression model for information loss prediction is trained. 
\item Sampling Policy Determination: When new data is sensed, the algorithm computes the current (last known) latent context and choses the best sampling policy given that context. The best policy is the one that minimizes the objective function which considers both sampling cost and predicted information loss.

\end{enumerate}
The suggested method is based on several hypotheses. The first is that different users have different behaviors which are reflected in sensors' data values and their derived contexts. For example, some users tend to hold their phone in their hands while others carry it in their pockets; some people perform activities such as eating while holding or touching their phone; and others seldom touch their phone \cite{hoober13}. Thus, building a personal model for context detection will result in a more accurate context than using a single model for all users. The second hypothesis is that the determination of the sampling policy should be done dynamically and take the user's most recent known context into consideration rather than an outdated or predefined set of known contexts. We believe that different contexts require different sampling policies. Different contexts may vary in their duration and in the list of sensors that are needed for their detection. For example, when entering an office, there is no point in sensing the GPS frequently, or when the user sleeps, the accelerometer data may be useless. Since our method can handle latent contexts that are not predefined or outdated, the dynamic nature of the detection is crucial. Finally, we hypothesize that Kullback-Leibler Divergence (KL-Divergence) is an applicable loss function for calculating the difference between the actual and the last known context vectors. In contrast to other difference metrics, KL-Divergence measures the difference between two non-symmetric probabilities vectors. While the first vector represents the "true" distribution, the second represents an approximation of that distribution. Therefore, when creating the first context vector based on actual sensors' data and the second context vector based on the last known sensors' data, the KL-Divergence between them reflects the information loss when reducing sensing cost. 
\vspace{-4pt}
\subsection{Data Extension}

In this step, we wish to simulate a situation of non-sampling sensors for a time period by extending each record to multiple synthetic records. A synthetic record contains previous sensors' values instead of actual values, as if they weren't sensed in that time interval.

In the extension algorithm, Algorithm \ref{AlgorithmExtension}, we iterate through all of the user's records (line 2). From each record we create multiple synthetic records by completing sensors' values from older records. The choice of which older record to use is derived from a specific distance $"dist"$ which defines the record's distance from the older record and varies from 1 to a predetermined configurable maximum limit $maxDist$. The distance is determined for each sensor separately and is maintained in a dedicated distances vector $\overrightarrow{D}$.

First, we add the actual record to the extended records list (line 6), thus the distance vector $\overrightarrow{D}$ will be all zeros (lines 4-5).
Second, we create synthetic records either systematically or randomly. When extending data systematically, we iterate on all possible distances in a range (line 7) and collect all sensors' values from a record which is located $"dist"$-records away (lines 12-14). In that case, the distance vector $\overrightarrow{D}$ will be all $"dist"$ (lines 10-11). When extending data randomly, for each sensor (line 17) we choose a random distance from a range, add it to the distances vector $\overrightarrow{D}$ (lines 18-19), and collect the sensors' values from the corresponding older record (lines 20-22). In other words, the values of features for different sensors are taken from different records. The process of creating a random record is repeated $k$ times (line 15).

\begin{algorithm}
\caption{Data Extension}\label{AlgorithmExtension}
\begin{flushleft}
\textbf{Input:} 
$records$ - User's records, 
$k$ - Number of random synthetic records per one actual record, 
$sensors$ - List of sensors, 
$maxDist$ - Maximal distance,
$features$ - List of features \linebreak
\textbf{Output:} $extendedRecords$ - Extended records list 
\end{flushleft}
\begin{algorithmic}[1]
\State Initialize $index$ and $extededRecords$
\ForAll{$r \in records$} \Comment{Iterate user's records}
\State $\overrightarrow{D} \gets$ initialize vector
\ForAll{$s \in sensors$}
\State $\overrightarrow{D}$.add(0)
\EndFor
\State $extendedRecords$ .add(r,$\overrightarrow{D}$)\Comment{maintain actual record}
\For{$(dist \gets 1$ to $maxDist)$} 
\State Initialize $\overrightarrow{D}$ and $newRecord$
\ForAll{$s \in sensors$}
\State $\overrightarrow{D}$ .add(dist)
\EndFor
\State $oldRecord \gets records[index+dist]$
\ForAll{$f \in features$}
\State $newRecord$.add($oldRecord[f]$)
\EndFor
\State $extendedRecords$.add($newRecord,\overrightarrow{D}$)
\EndFor
\For{$i \gets$ 1 to k}
\State Initialize $\overrightarrow{D}$ and $newRecord$
\ForAll{$s \in sensors$}
\State $dist \gets random(1,maxDist)$
\State $\overrightarrow{D}$ .add(dist)
\State $oldRecord \gets records[index+dist]$
\ForAll{$f \in features[sensor]$}
\State $newRecord$.add($olderRecord[f]$)
\EndFor
\EndFor
\State $extendedRecords$.add($newRecord,\overrightarrow{D}$)
\EndFor
\State $index \gets index+1$
\EndFor
\State return $extendedRecords$
\end{algorithmic}
\end{algorithm}

%Figure \ref{dataExtensionImage} demonstrates the extension of a single record, first systematically and then randomly. The colors reflect the records in the original dataset from which the sensors' values are taken for the synthetic records. The difference between systematic and random extension can be clearly observed. In systematic extension block all values are taken from the same record, thus each row is colored with only one color. For the random extension each sensor obtains a different color, since each sensor has a different distance. 

\subsection{Latent Context Detection}
In the Latent Context Detection step, we build a personal model for latent context detection for each user. 

Latent contexts are hidden context patterns modeled as numeric vectors. They can be obtained automatically by applying unsupervised learning techniques on available raw data \cite{unger16}. Our work is the first to create an energy efficient framework for latent context detection, and therefore it is not limited to a predefined set of contexts.

\begin{figure}
\centering
\includegraphics[width=200px,height=\linewidth,keepaspectratio]{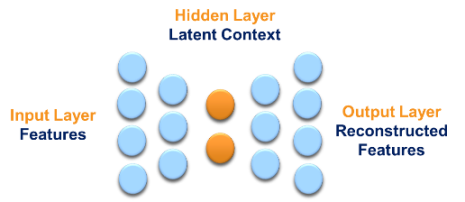}
\caption{Personal autoencoder model for latent context detection}\label{LatentContextImage}
\end{figure}

We chose to use Unger's \cite{unger16} method for latent context detection using an autoencoder. The autoencoder is a neural network model aiming at reconstructing the input after reducing the input's dimension by adapting the autoencoder's weights. When defining the input as sensors' values (as can be seen in Figure \ref{LatentContextImage}), we consider the most hidden layer as the user's latent context. In other words, the latent context vector is a reduced dimensional representation of the features' vector. Since our goal is to predict the hidden layer, the output layer is only used for training the model and can be discarded afterwards.

Algorithm \ref{AlgorithmLatentContext} describes the process of latent context detection. After creating a personal context detection model (line 1), we iterate all of the user's records (line 3) and detect latent context for each record (line 4). The context $\overrightarrow{C}$ is saved, along with the distances $\overrightarrow{D}$ from the previous step (line 6). The sensors' values are no longer required; only the context and the distances between samples are needed.

\begin{algorithm}
\caption{Latent Context Detection}\label{AlgorithmLatentContext}
\begin{flushleft}
\textbf{Input:} $records$ - User's records,
$extendedRecords$ - User's extended records \linebreak
\textbf{Output:} $contextDistancesRecords$ - List of context and distances for each extended record
\end{flushleft}
\begin{algorithmic}[1]
\State $model \gets CreateAutoEncoderModel(records)$
\State $contextDistancesRecords \gets$ initialize list
\ForAll{$r \in extendedRecords$}
\State $\overrightarrow{C} \gets getContext(model,r)$
\State $\overrightarrow{D} \gets getDistances(r)$
\State $contextDistancesRecords$.add({$\overrightarrow{C},\overrightarrow{D}$})
\EndFor
\State return $contextDistancesRecords$
\end{algorithmic}
\end{algorithm}

\subsection{Information Loss Prediction}

During the Information Loss Prediction step, our goal is learning the information loss when sampling sensors in different frequencies (i.e., distances between samples). We consider the information loss as the distance between the synthetic record's context and its actual record's context. In other words, the change in sensors' values results in different context, and the distance between the two contexts is the information loss.

We chose to calculate this distance using KL-Divergence, which is a measure of the difference between two non-symmetric probability distributions $P$ and $Q$. $P$ typically represents the "true" distribution of data, while $Q$ typically represents an approximation of $P$ \cite{Burnham03}. Therefore, in our algorithm, $P$ is the actual record's context (the "true" context), while $Q$ is the artificial record's context. The formula for calculating KL-Divergence is:
\begin{equation}\label{eqKL}
\int_{-\infty}^{\infty}p(x)log_{2}\frac{p(x)}{q(x)} dx
\end{equation}

Algorithm \ref{AlgorithmInformationLoss}, which is used to calculate and predict information loss, is described next. Given the output from Algorithm \ref{AlgorithmLatentContext}, for each record (synthetic or not), we extract the context and distances (lines 3-4) and retrieve the actual record's context (line 5). The actual record can be found by a simple calculation based on the current record's index (since we know how many synthetic records have been created for each record). Afterwards, using Eq. \ref{eqKL}, we calculate the information loss (line 6) and create a new record that contains the actual context, the corresponding distances, and the derived information loss (line 7).

\begin{algorithm}
\caption{Information Loss Prediction}\label{AlgorithmInformationLoss}
\begin{flushleft}
\textbf{Input:} 
$contextDistances$ - List of contexts and corresponding distances,
$contextModel$ - Latent Context detection model\linebreak 
\textbf{Output:} $infoLossModel$ - Lasso model for information loss prediction
\end{flushleft}
\begin{algorithmic}[1]
%\State $index \gets 0$
%\State $ContextDistancesInfoLoss \gets$ initialize list
\State Initialize $ContextDistInfoLoss$ and $index$ 
\ForAll{$r \in contextDistances$}
\State $\overrightarrow{D} \gets getDistances(r)$
\State $\overrightarrow{C} \gets getContext(contextModel,r)$
\State $\overrightarrow{AC} \gets getActualContext(index)$
\State $infoLoss \gets calculateInfoLoss(\overrightarrow{AC},\overrightarrow{C})$
\State $ContextDistInfoLoss$.add({$\overrightarrow{AC},\overrightarrow{D},infoLoss$})
\State $index \gets index+1$
\EndFor
\State $transformedData \gets tranform(ContextDistInfoLoss)$
\State $infoLossModel \gets trainModel(transformedData)$
\State return $infoLossModel$
\end{algorithmic}
\end{algorithm}
After calculating the KL-Divergence between each pair of the synthetic record's context and its actual record's context, we build a personal model for predicting information loss given the last known context and a set of distances (one distance for each sensor). We chose to use a linear regression model with KL-Divergence as its dependent variable. The independent variables are the last known context $\overrightarrow{C}$, the distances from last sample $\overrightarrow{D}$, their squared values, and the interaction variables between them, so that the decision will be dependent on the context. These variables form the features of the regression model.

Since the context is set when determining a sampling policy, without interaction variables it is treated as a constant and is ignored. Therefore, in line 9 we transform $contextDistInfoLoss$ so it also contains the squared values and interaction variables. The information loss is predicted according to the following regression formula:

\begin{equation}\label{eqRegModel}
\begin{split}
infoLoss(\overrightarrow{C},\overrightarrow{D}) & = \sum_{i=1}^{n}{b_c}_i C_i+\sum_{j=1}^{m}{b_d}_j D_j + \\& \sum_{i=1}^{n}{b_{sc}}_i C_i^2+\sum_{j=1}^{m}{b_{sd}}_j D_j^2+\sum_{i=1}^{n}\sum_{j=1}^{m}{b_{c_{i}d_{j}}} C_i D_j
\end{split}
\end{equation}
where $1\leqslant i \leqslant|C|=n$, $1\leqslant j \leqslant|D|=m$, $b_c$ and $b_d$ are context and distance coefficients, $b_{sc}$ and $b_{sd}$ are squared value coefficients, and $b_{cd}$ are interaction variable coefficients.

After creating the features, we train a Lasso model (line 10) and return it (line 11).
Our first choice for the information loss prediction model was the state of the art $XGBoost$ model for regression. When comparing it to a simple linear regression model, we found that although the $XGBoost$ error is lower, the difference in the MSE between the models is not statistically significant. However, we preferred to use the pure regression model, since it is a much simpler model. Furthermore, in order to force positive coefficients, we used a Lasso model which reduced the model's complexity by setting some of the coefficients to zero.

\subsection{Policy Determination}

For this step, which is the final step in the process, we use the results of all of the prior steps. When new data is sensed, we wish to determine when to resample each sensor. A decision about the sampling policy is made based on the sensors' sampling cost, predicted information loss, and the last known context.

Therefore, when new data is sensed, the following process is employed. First, we detect the current context using the personal model we built in the Latent Context Detection step. Then, given that context, we choose the best policy (i.e., best distance) for each sensor. The best policy is the one that minimizes the sampling cost while incurring minimal information loss. Thus, the determining of sampling policy is based on the following functions:
\begin{description}
\item$\bullet$ Cost function: This function considers the sampling cost of each sensor and the sampling frequency (distance from last sample). It puts the cost in direct proportion to the sensors' costs and in inverse proportion to the distances between samples. Where $n$ is number of sensors and $\overrightarrow{D}$ is distances vector, the cost is computed using the following formula: 
\begin{equation}\label{eqCost}
cost(\overrightarrow{D})=\sum_{i=1}^{n}\frac{cost_i}{D_i}
\end{equation}
\item$\bullet$ Information loss function: A Lasso regression model, as seen in Eq. \ref{eqRegModel}, with KL-Divergence as its dependent variable. Independent variables are derived from the last known context $\overrightarrow{C}$ and distances between samples $\overrightarrow{D}$.

\end{description}
The objective function will include both sampling cost and information loss:%, and will be solved by $cvxopt$ package in Python:
\begin{equation}\label{eqObjective}
\min\left(\sum_{i=1}^{n}\frac{cost_i}{D_i}+\alpha \times infoLoss(\overrightarrow{C},\overrightarrow{D})\right)
\end{equation}	
where $n$ is the number of sensors, $|\overrightarrow{D}|=n$, and $\alpha$ is a tuning parameter for the weight of the information loss. 
%\begin{description}
%\item$\bullet$ $n$ - Number of sensors
%\item$\bullet$ $|\overrightarrow{D}|=n$
%\item$\bullet$ $\alpha$ - A tuning parameter for the weight of the information loss 
%\end{description}

After determining the sampling policy, we set a countdown timer which repeatedly triggers the policy determination. It raises the questions of when should the sampling policy be determined again or what value should we initialize the timer with. We considered the following options: (1) MAX: The maximal distance from the last policy; (2) MIN: The minimal distance from the last policy; (3) AVG: The average distance from the last policy; (4) NEVER: determine the policy once and never determine it again. For example: If we have three sensors and the last policy was 1,5,6, then according to MAX, the next policy will be determined after six time intervals, according to AVG after four intervals, and according to MIN after one interval. However, NEVER will initialize the timer with infinity. In the next section, we evaluate the difference between those methods with respect to information loss and cost trade-offs. Switching between them can be handled by changing a configurable parameter. 

The complete algorithm is described in Algorithm \ref{AlgorithmPolicyDetermination}. It gets as input a list of sensors, policy determination mode and user's personal models for information loss prediction and context detection. First, we initialize $timeToSample$ and $policyTimer$ with zeros (lines 1-4). The first indicates how many time intervals are left to re-sample each sensor, and the second indicates how much time intervals are left to switch policy. We initialize them with zeros in order to sample all sensors and determine a sampling policy the minute the process starts. 

\begin{algorithm}
\caption{Continuous Policy Determination}\label{AlgorithmPolicyDetermination}
\begin{flushleft}
\textbf{Input:} 
$infoLossModel$ - Regression model for predicting information loss, 
$contextModel$ - Autoencoder model for context detection,
$sensors$ - List of sensors,
$mode$ - "MIN", "MAX", "AVG" or "NEVER",
$maxDist$ - Maximal distance
\end{flushleft}
\begin{algorithmic}[1]
\State $timeToSample \gets$ initialize dictionary
\ForAll{$s \in sensors$}
\State $timeToSample[s] \gets 0$
\EndFor
\State $policyTimer \gets 0$
%\State $prevPolicy \gets null$
%\State $record \gets$ initialize list
\State Initialize $policyTimer$, $prevPolicy$ and $record$
\While{true}
\ForAll{$s \in sensors$}
\If{$timeToSample[s]=0$}
\State $lastSampledValues[s] \gets sampleSensor()$ 
\EndIf
\State record.add($lastSampledValues[s]$) 
\EndFor
\State $context \gets calculateContext(contextModel,record)$
\If{$policyTimer =0$} 
\State $prevPolicy \gets policy$
\State $policy \gets$ Determine policy with Algorithm \ref{AlgorithmOptimization}
\State $timeToSample \gets$ Get value from Algorithm \ref{AlgorithmTimeLeft}
\State $policyTimer\gets setPolicyTimer(mode,policy)$
\Else
\ForAll{$s \in sensors$}
\State $timeToSample[s]\gets timeToSample[s]-1 $
\If{$timeToSample[sensor]< 0$}
\State $timeToSample[s] \gets policy[s]$
\EndIf
\EndFor
\State $policyTimer \gets timeToPolicy-1$
\EndIf
\EndWhile
\end{algorithmic}
\end{algorithm}

After initialization, we perform an infinite loop and repeat the following process: First, for each sensor we check if next sampling countdown has reached zero (line 8). In case it has, we sense it and save its new values in $lastSampleValues$ dictionary (line 9). Otherwise, we do nothing and last sensed values are kept. Either way, we append sensor's values (new or old) to $record$, which represents current user's record (line 10). Second, we detect current context using user's context model (line 11). Third, if time to new policy has reached zero (line 12), we save current policy as $prevPolicy$ (line 13), determine a new policy using Algorithm \ref{AlgorithmOptimization} (line 14), recalculate $timeToSample$ for all sensors using Algorithm \ref{AlgorithmTimeLeft} (line 15) and update $policyTimer$ according to the given mode (line 16). Otherwise, we simply reduce by one $timeToSample$ of all sensors (line 18-19), and $timeToPolicy$ as well (line 22). If $timeToSample$ of one of the sensors is negative, we initialize it with the value from current policy (lines 20-21).

Algorithm \ref{AlgorithmOptimization} describes the use of $cvxsolver$ (from Python's $cvxopt$ package) in order to minimize objective function. In lines 1-2 we create $G$ and $h$, which together represent a constraints system, such that $Gx\leqslant h$. $G$ is a sparse matrix of size $2n\times n$, where value $(i,j)$ represents the coefficient of variable $j$ in equation $i$. $h$ is a dense matrix of size $2n\times 1$, where value $(i,1)$ represents the right-hand side constant for equation $i$. Since we would like to achieve $1\leqslant x\leqslant maxDist$ for every distance $x$, we initialize $G$ and $h$ in the following way: First $n$ rows of $G$'s diagonal and first $n$ rows of $h$ are initialized with -1 and 1 respectively, while last $n$ rows of $G$'s diagonal and last $n$ rows of $h$ are initialized with 1 and $maxDist$ respectively. After defining the constraints, we define the target function $F$ with Eq. \ref{eqObjective} (line 3) and call the optimization solver with $F$,$G$ and $h$ (line 4). At last, best policy is returned (line 5).

\begin{algorithm}
\caption{Optimize Policy}\label{AlgorithmOptimization}
\begin{flushleft}
\textbf{Input:}
$infoLossModel$ - Regression model for predicting information loss, 
$context$ - User's last known context, 
$n$ - Number of sensors,
$maxDist$ - Maximal distance\linebreak
\textbf{Output:} 
$policy$ - Optimized sampling policy
\end{flushleft}
\begin{algorithmic}[1]
\State $G \gets createSparseMatrix(n)$
\State $h \gets createDenseMatrix(n,1,maxDist)$
\State $F \gets$ objective function as described in Eq. \ref{eqObjective}
\State $policy \gets cvxsolver(F,G,h)$ 
\State return policy
\end{algorithmic}
\end{algorithm}

\begin{algorithm}
\caption{Determine Time Left to Sample}\label{AlgorithmTimeLeft}
\begin{flushleft}
\textbf{Input:} 
$policy$ - New selected policy, 
$prevPolicy$ - Previous policy,
$timeToSample$ - Number of time intervals that are left until the next sampling for each sensor,
$sensors$ - List of sensors\linebreak
\textbf{Output:} 
$timeToSample$ - Updated number of time intervals that are left until the next sampling for each sensor
\end{flushleft}
\begin{algorithmic}[1]
\If{$prevPolicy \neq null$}
\If{$prevPolicy \neq policcy$}
\ForAll{$s \in sensors$}
\If{$timeToSample[s]=0$}
\State $timeToSample[s]\gets policy[s] $
\Else
\State $timeSinceSampled\gets prevPolicy[s]-$ $timeToSample[s]$
\State $time\gets policy[s]-timeSinceSampled$
\State $timeToSample[s] \gets max(0,time)$
\EndIf
\EndFor
\EndIf
\Else
\State $timeToSample \gets policy$
\EndIf 
\State return $timeToSample$
\end{algorithmic}
\end{algorithm}

Algorithm \ref{AlgorithmTimeLeft} describes the way we recalculate $timeLeft$-$ToSample$ for each sensor according to the new policy. If previous policy is $null$, it means that this is the first time we determine a sampling policy and therefore $timeToSample$ will be equal to current policy (line 11). Otherwise, we need to take into consideration the previous policy. If previous policy is equal to new policy, we do nothing and $timeToSample$ stays the same. If not (line 2), we do the following for each sensor: If $timeToSample$ of a sensor is zero, we simply initialize it with its new policy value (lines 4-5). Otherwise, we first calculate the number of time intervals that have passed since last sample ,$timeSinceSampled$, by subtracting $timeToSample$ of the sensor from its previous policy value (line 7). Second, we calculate the remaining time according to the new policy by subtracting $timeSinceSampled$ from its new policy value (line 8). In case of negative result, $timeLeftToSample$ is set to zero (line 9).

\section{Evaluation}

In this section, we describe the evaluation of our method. The objective of the evaluation was to show that it is beneficial to dynamically determine the sampling policy in terms of accuracy and energy consumption. In addition, we wanted to show that KL-Divergence is an applicable information loss measure for latent context detection.

We performed a series of offline simulations on the Sherlock dataset \cite{Mirski16}. We used data collected from twenty users from six sensors of their mobile devices, namely: GPS, cell tower, accelerometer, gyroscope, magnetic field, and status (including various status features of the phone, such as volume, screen orientation, etc.). All sensors were sampled once a minute. The data was collected for about a week and contains about 10,000 records for each user. 

\begin{figure}
\centering
\includegraphics[width=200px,height=\linewidth,keepaspectratio]{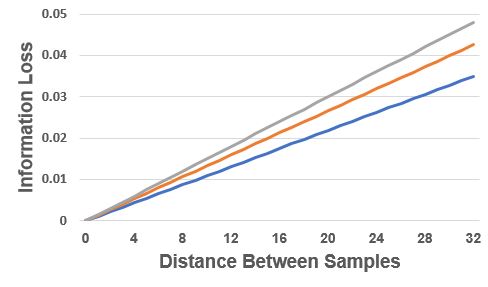}
\caption{Predicting information loss for three different contexts (each line represent different context)}\label{predKLImage}
\end{figure}

We use KL-Divergence as a measure for information loss between the actual and last known contexts. In order to check its applicability as a measure for information loss, we trained twenty personal models for information loss prediction, one for each user, and checked its correlation with the distances between samples across different contexts.

After an empirical evaluation, we set $maxDist=32$ and $K=20$ when extending the user's data (the first step of our method), meaning that from a single record we simulated 52 synthetic records. 32 were generated using the systematic method, and 20 were generated randomly. We then calculated the KL-Divergence between their derived contexts. For all users, the regression model for the information loss prediction succeeded in converging to a solution, when considering the positive coefficient constraint and taking some of the context features into account (i.e., some context features resulted in non-zero coefficients). Then, for each user we predicted the information loss when setting the same distance for all sensors from 1 to $maxDist$. Figure \ref{predKLImage} presents the information loss as a function of the distance between samples for a single user, and each linear line stands for a different context. It shows that the greater the distance between samples, the greater the predicted information loss. Moreover, it can been seen that for the same user, the slope of the graph varies for different contexts. This indicates that the last known context affects the predicted information loss and thus may affect the sampling policy.

\subsection{Energy-Information Loss Trade-off}

During the policy determination step, we simulated the process of continuous dynamic sensing that includes iterations of the following: a)
Sample sensors according to the determined policy; b) Detect context; c) When required - determine new policy.

The selected policy in each iteration is the one that minimizes the sum of cost (energy) and information loss given the last known context (See Eq. \ref{eqObjective}). The process is repeated continuously, calculating the total information loss and the total cost for all user's records. 

In the continuous process of sensing and determining sampling policy, the question of when to redetermine policy was raised. We simulated four different methods: (1) MAX: The maximal distance from last policy; (2) MIN: The minimal distance from last policy; (3) AVG: The average distance from last policy; (4) NEVER: determine policy once and never determine it again. The NEVER method represents the static method. Each method was tested with five different weights for information loss (See Eq. \ref{eqObjective}): 0.1, 1, 5, 10 and 20. Hence, in total, we performed twenty simulations for each user.

The process of ranking the methods goes as follows: Each simulation is represented as a two-dimensional point $(x,y)$, where x and y represent the total information loss and total cost respectively. Then, for each $\alpha$, we find a subset of points that constitutes the Pareto Frontier (for minimum cost and minimum information loss), subtract it from the superset, and repeat the process until the superset is empty. With each iteration, the rank is increased by one. After rank calculation, we computed the statistic $F_F=29.492$, which has turned out to be greater than the critical value (2.696). Therefore, the null hypothesis was rejected. Table \ref{tblRank} presents the average rank for each method. Results show that the dynamic methods perform better than the static method (NEVER), and the more frequent the decision, the better the rank. This implies that frequent policy determination improves results.

\begin{figure}
\centering
\includegraphics[width=240px,height=\linewidth,keepaspectratio]{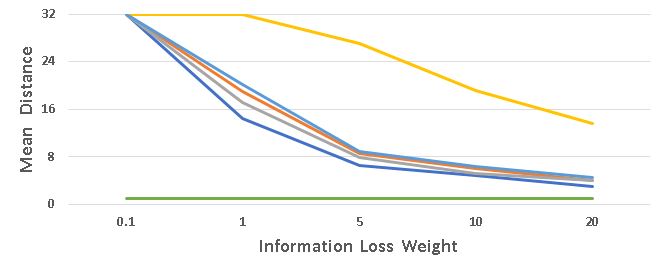}
\includegraphics[width=240px,height=\linewidth,keepaspectratio]{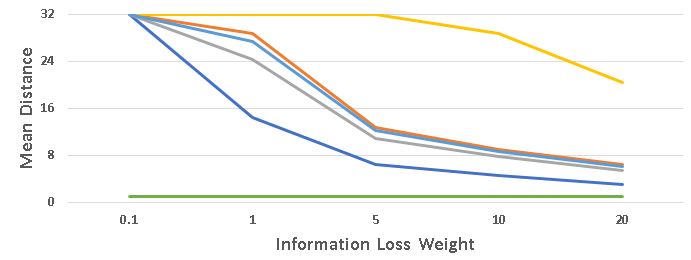}
\includegraphics[width=240px,height=\linewidth,keepaspectratio]{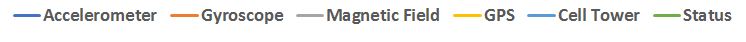}
\caption{Mean distance between samples as a function of information loss weight}\label{AvgDistImage}
\end{figure}

\iffalse
\begin{table}[ht]
\caption{Critical values for two-tailed Nemenyi test}
\label{tblCritical}
\centering 
\noindent\adjustbox{max width=\linewidth}{
\begin{tabular}{c | c c c c c c c c c} 
\hline 
\#Classifiers & 2 & 3 & 4 & 5 & 6 & 7 & 8 & 9 & 10\\ [0.5ex] % 
\hline 
$q_{0.05}$&1.960&2.343&2.569&2.728&2.850&2.949&3.031&3.102&3.164\\
$q_{0.10}$&1.645&2.052&\cellcolor{gray!20}2.291&2.459&2.589&2.693&2.780&2.855&2.920\\[1ex]
\hline 
\end{tabular}
}
\end{table}
\fi

\begin{table}[ht]
\caption{Mean rank of policy determination timing methods} 
\label{tblRank}
\centering 
\begin{tabular}{c | c c c c} 
\hline 
Method & MIN & AVG & MAX & NEVER \\ [0.5ex] 
\hline 
Mean Rank & 1.685 & 2.360 & 2.915 & 3.040 \\ 
\hline
\end{tabular}
\label{table:nonlin} 
\end{table}

After rejecting the null hypothesis, we performed the Nemenyi test between each pair of methods. The conclusion is that there's a significant difference between all methods except for NEVER and MAX. We assume this is due to the fact that policy doesn't change for a relatively long time window, whereas many changes in context may occur. In addition, as we expected, when examining optimized sampling policies for all sensors, we saw that the greater the $\alpha$ (information loss weight), the smaller the distances between samples. The only exception is the status sensor, which constantly gets policy of one (sample every time interval). This makes sense, since it's the only sensor which has zero cost. However, the GPS sensor, which is the most expensive sensor, gets $maxDist$ policy even with $\alpha$ greater than 0.1. Figure \ref{AvgDistImage} presents the average policy for each sensor as a function of $\alpha$ for two users. Each line refers to a different sensor. It can been seen that for all sensors, except for the status sensor, the average distance between samples decreases when the weight of information loss increases. This is clear evidence that our method successfully performs an energy-information loss trade-off that can be controlled by applying different weights.

\subsection{Comparison with State-of-the-Art}
We chose to compare our method with Rachuri's \cite{rachuri11} approach which is the only method to the best of our knowledge that uses machine learning to perform adaptive sampling.
Rachuri \cite{rachuri11} suggests to adjust the sensor's duty cycle according to its sensing probability. The sensing probability is dynamically calculated and defined as the portion of successes of previous sensing actions. A success indicates that the sensing action resulted in capturing an interesting event (using domain dependent pretrained classifier). The sensors are sampled at a high rate when interesting events are observed and at a low rate when there are no events of interest. The technique works as follows: Let $p_i$ be the probability of sensing from a sensor $s_i$ where $i \in {accelerometer, Bluetooth, microphone}$, and $a_i$ is the sensing action on a sensor $s_i$. If the sensing action $a_i$ results in an interesting event, the probability is increased: $p_i = p_i + \alpha (1 - p_i)$, where $0 < \alpha < 1$. Otherwise, the probability is decreased: $p_i = p_i - \alpha p_i$. The lower and upper bounds of the probabilities were limited to 0.1 and 0.9, respectively. While his approach considers the values of each sensor separately, our approach provides a more complete view of a user's context by considering the combination of multiple sensors.

\begin{figure}[H]
\centering
\includegraphics[width=200px,height=\linewidth,keepaspectratio]{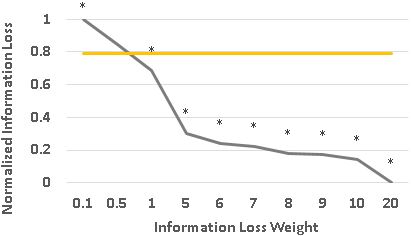}
\includegraphics[width=200px,height=\linewidth,keepaspectratio]{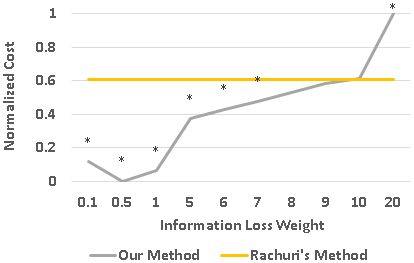}
\caption{Normalized mean information loss and mean cost as a function of information loss weight}\label{avgRessults}
\end{figure}

In order to compare our method to Rachuri's method, we implemented it with a few necessary adjustments: First, since our data is not labeled we weren't able to train event classifiers. Therefore, in order to determine whether an interesting event has occurred, we calculated the KL-Divergence between every pair of consecutive sensor records and used the 90\% quantile as a threshold. Second, since we use offline simulations we could only change the time between samples and weren't able to change the duty cycle. Therefore, according to the adapted sensing probability, we calculated the time between samples such that lower probability results in longer time between samples.

We ran offline simulations  using our implementation for each of the users and calculated the total cost and information loss; and compared the results with our values when using the MIN timing approach over different information loss weights. To determine the significance of the difference between the performance of methods we used a paired t-test with $\alpha = 0.01$ as the confidence level. The results are presented in Figure \ref{avgRessults} which provides a comparison of Rachuri's normalized mean information loss and cost with our normalized mean values as a function of the weight. The statistically significant results are denoted by an asterisk (*). The results demonstrate that in terms of information loss, our method is significantly better when the weight is 1 or higher, and in terms of cost, it is significantly better when the weight is 7 or lower. Therefore, our method is better for both cost and information loss when the weight is in the range of 1 to 7.
%We compared our MIN approach against Rachuri's method used our method with the MIN approach and the same information loss weights as we used in previous experiments (0.1, 1, 5, 10, 20). For each weight, we calculated the total cost and total information loss and compared these values with those of Rachuri's method using a paired t-test with $\alpha = 0.01$ as the confidence level. The computed statistics are presented in Table \ref{tblTtest}.The results where our method performed better are denoted in bold, while results that are statistically significant are denoted by an asterisk (*). Results show that in terms of information loss our method is significantly better for weights 1 or higher, and it terms of cost it is significantly better for weights 7 or lower. Therefore, our method is better in both cost and information loss for weights between 1 to 7 (denoted in gray).
%
\iffalse
\begin{table}[ht]
\caption{Paired T-test results} 
\label{tblTtest}
\centering 
\begin{tabular}{ c | c  c } 
\hline 
Weight & Information Loss &
Cost \\
\hline
0.1 & 4.089* & \bf 18.874* \\
0.5 & 1.425 & \bf 19.734*\\
1 & \cellcolor{gray!20} \bf 2.726* &  \bf \cellcolor{gray!20} 17.414* \\
5  & \cellcolor{gray!20} \bf 7.560* & \cellcolor{gray!20} \bf 8.304 \\
10 & \bf 8.189* & 0.212 \\
20 & \bf 9.010*  &  5.986* \\ [0.5ex] 
\hline
\end{tabular}
\label{table:nonlin} 
\end{table}
\fi

\section{Conclusions and Future Work}

We presented a novel method for continuous cost-aware sensing for managing the trade-off between energy consumption and information loss while trying to minimize both. The suggested framework dynamically determines a user's sampling policy based on three factors: (1) User's current latent context; (2) Predicted information loss; (3) Sensors' sampling costs. The latent context is calculated with an autoencoder, the information loss is predicted with Lasso regression, and the best sampling policy is determined using convex optimization. All models are personal, and objective function is a nonlinear function that gives weight to both sampling costs and predicted information- loss. We evaluated the suggested method by performing a series of offline simulations on data recorded from six mobile device sensors of twenty users. 

The results show that the dynamic adaptation approach is better than the static approach in terms of accuracy and energy consumption, and that KL-Divergence is an applicable measure for information loss for the task of latent context detection. 

The results show that our method successfully performs an energy-information loss trade-off that can be controlled by setting different weights in the objective function. This enables context-aware applications to be accurate while consuming less energy.

Moreover, when comparing our method to another state of the art dynamic method, it outperformed in both the sampling cost and information loss measures in some cases, while in other cases our method achieved better results in one of those measures.

In the future, we plan to implement and evaluate our method within a context-aware application and compare the performances of personal vs non-personal models. In addition, we plan to improve our method by adding feature selection on users' sensors data. This is due to the assumption that some features may perform better than others for different users. Furthermore, we wish to use hybrid models which optimize the sampling policy based on both personal and non-personal models. The hybrid models solve the cold start problem when there is insufficient user data or when model's error exceeds a predetermined threshold.

\clearpage
\bibliographystyle{ACM-Reference-Format}

\begin{thebibliography}{11}
\bibitem{ABOWD99}		Abowd, Gregory D., et al.\textit{"Towards a better understanding of context and context-awareness"}, In: International Symposium on Handheld and Ubiquitous Computing. Springer Berlin Heidelberg, 1999. p. 304-307.
\bibitem{Muhtadi03}		Al-Muhtadi, Jalal, et al. \textit{"Cerberus: a context-aware security scheme for smart spaces."} Pervasive Computing and Communications, 2003.(PerCom 2003). Proceedings of the First IEEE International Conference on. IEEE, 2003.
\bibitem{Alhamid16}		Alhamid, Mohammed F., et al. \textit{"RecAm: a collaborative context-aware framework for multimedia recommendations in an ambient intelligence environment."} Multimedia Systems 22.5 (2016): 587-601.
\bibitem{BALDAUF07}		Baldauf, Matthias; Dustdar, Schahram; Rosenberg, Florian.\textit{"A survey on context-aware systems"}, International Journal of Ad Hoc and Ubiquitous Computing, 2007, 2.4: 263-277.
\bibitem{ben09}		 	Ben Abdesslem, Fehmi; Phillips, Andrew; Henderson, Tristan. \textit{"Less is more: energy-efficient mobile sensing with senseles"}, In: Proceedings of the 1st ACM workshop on Networking, systems, and applications for mobile handhelds. ACM, 2009. p. 61-62.
\bibitem{Linas11}		Baltrunas, Linas, et al.\textit{"Incarmusic: Context-aware music recommendations in a car."} E-Commerce and web technologies (2011): 89-100.
\bibitem{Bardram04}		Bardram, Jakob E. \textit{"Applications of context-aware computing in hospital work: examples and design principles."} Proceedings of the 2004 ACM symposium on Applied computing. ACM, 2004.
\bibitem{Bardram03}		Bardram, Jakob E. \textit{"Hospitals of the future–ubiquitous computing support for medical work in hospitals."} Proceedings of UbiHealth. Vol. 3. 2003.
\bibitem{Burnham03}		Burnham, Kenneth P., and David Anderson. \textit{"Model selection and multi-model inference."} A Pratical informatio-theoric approch. Sringer 1229 (2003).
\bibitem{Chakraborty13}		Chakraborty, Supriyo, et al. \textit{"A framework for context-aware privacy of sensor data on mobile systems."} Proceedings of the 14th Workshop on Mobile Computing Systems and Applications. ACM, 2013.
\bibitem{Chen00}		Chen, Guanling, and David Kotz. \textit{"survey of context-aware mobile computing research."} Vol. 1. No. 2.1. Technical Report TR2000-381, Dept. of Computer Science, Dartmouth College, 2000.
\bibitem{Janez06}		Demsar, Janez \textit{"Statistical comparisons of classifiers over multiple data sets."}, Journal of Machine learning research 7.Jan (2006): 1-30.
\bibitem{Friedman37}		Friedman, Milton. \textit{"The use of ranks to avoid the assumption of normality implicit in the analysis of variance."} Journal of the american statistical association 32.200 (1937): 675-701.
\bibitem{Friedman40}		Friedman, Milton.\textit{"comparison of alternative tests of significance for the problem of m rankings."} The Annals of Mathematical Statistics 11.1 (1940): 86-92.
\bibitem{hoober13}		Hoober, Steven. \textit {"How do users really hold mobile devices."} Uxmatters (http://www. uxmatter. com). Published: Feburary 18 (2013).
\bibitem{Iman80}		Iman, Ronald L., and James M. Davenport.\textit{"Approximations of the critical region of the fbietkan statistic."} Communications in Statistics-Theory and Methods 9.6 (1980): 571-595.
\bibitem{kabir15}		Kabir, M. Humayun, M. Robiul Hoque, and Sung-Hyun Yang. \textit{"Development of a Smart Home Context-aware Application: A Machine Learning based Approach."} learning 9.1 (2015).
\bibitem{Kang06}		Kang, Dong-Oh, et al. \textit{"A wearable context aware system for ubiquitous healthcare."} Engineering in Medicine and Biology Society, 2006. EMBS'06. 28th Annual International Conference of the IEEE. IEEE, 2006.
\bibitem{kang08}		Kang, Seungwoo, et al. \textit{"Seemon: scalable and energy-efficient context monitoring framework for sensor-rich mobile environments."} Proceedings of the 6th international conference on Mobile systems, applications, and services. ACM, 2008.
\bibitem{Kansal13}		Kansal, Aman, et al.\textit{"The latency, accuracy, and battery (lab) abstraction: programmer productivity and energy efficiency for continuous mobile context sensing"}, ACM SIGPLAN Notices 48.10 (2013): 661-676.
\bibitem{Kjeldskov04}		Kjeldskov, Jesper, and Mikael Skov. \textit{"Supporting work activities in healthcare by mobile electronic patient records."} Computer Human Interaction. Springer Berlin/Heidelberg, 2004.
\bibitem{lockhart14}		Lockhart, Jeffrey W.; WEISS, Gary M. \textit{"The benefits of personalized smartphone based activity recognition models."} In: Proceedings of the 2014 SIAM international conference on data mining. Society for Industrial and Applied Mathematics, 2014. p. 614 622
\bibitem{Mirski16}		Mirsky Yisroel, Asaf Shabtia, Lior Rokach, Bracha Shapira, and Yuval Elovici, \textit{"SherLock vs Moriarty: A Smartphone Dataset for Cybersecurity Research"}, 9th ACM Workshop on Artificial Intelligence and Security (AISec) with the 23nd ACM Conference on Computer and Communications (CCS), 2016.
\bibitem{Mirsky17}		Mirsky, Yisroel, et al. \textit{"Anomaly detection for smartphone data streams."} Pervasive and Mobile Computing 35 (2017): 83-107.
\bibitem{Mitchell00}		Mitchell, Scott, et al. \textit{"Context-aware multimedia computing in the intelligent hospital."} Proceedings of the 9th workshop on ACM SIGOPS European workshop: beyond the PC: new challenges for the operating system. ACM, 2000.
\bibitem{Munuz03}		Muñoz, Miguel A., et al. \textit{"Context-aware mobile communication in hospitals."} Computer 36.9 (2003): 38-46.
\bibitem{Nath12}		Nath, Suman. \textit{"ACE: exploiting correlation for energy-efficient and continuous context sensing."} Proceedings of the 10th international conference on Mobile systems, applications, and services. ACM, 2012.
\bibitem{Nemenyi63}		P. B. Nemenyi. \textit{"Distribution-free multiple comparisons."} PhD thesis, Princeton University, 1963.
\bibitem{paek10}		Paek, Jeongyeup; KIM, Joongheon; GOVINDAN, Ramesh.\textit{"Energy-efficient rate-adaptive GPS-based positioning for smartphones"}, In: Proceedings of the 8th international conference on Mobile systems, applications, and services. ACM, 2010. p. 299-314.
\bibitem{Perera14}		Perera, Charith, et al.\textit{"Context aware computing for the internet of things: A survey"}. IEEE Communications Surveys \& Tutorials 16.1 (2014): 414-454.
\bibitem{rachuri10b}		Rachuri, Kiran K., et al.\textit{"EmotionSense: a mobile phones based adaptive platform for experimental social psychology research"}, In: Proceedings of the 12th ACM international conference on Ubiquitous computing. ACM, 2010. p. 281-290.
\bibitem{rachuri11}		Rachuri, Kiran K., et al.\textit{"Sociablesense: exploring the trade-offs of adaptive sampling and computation offloading for social sensing"},In: Proceedings of the 17th annual international conference on Mobile computing and networking. ACM, 2011. p. 73-84.
\bibitem{rachuri10a}		Rachuri, Kiran K.; MUSOLESI, Mirco; MASCOLO, Cecilia.\textit{"Energy-accuracy trade-offs in querying sensor data for continuous sensing mobile systems"}, In: Proc. of Mobile Context-Awareness Workshop. 2010.
\bibitem{sarker16}		Sarker, Sujan; NATH, Amit Kumar; RAZZAQUE, Abdur. \textit{"Tradeoffs between sensing quality and energy efficiency for context monitoring applications"}, In: Networking Systems and Security (NSysS), 2016 International Conference on. IEEE, 2016. p. 1-7.
\bibitem{Skubic09}		Skubic, Marjorie, et al. \textit{"A smart home application to eldercare: Current status and lessons learned."} Technology and Health Care 17.3 (2009): 183-201.
\bibitem{Solanas14}		Solanas, Agusti, et al. \textit{"Smart health: a context-aware health paradigm within smart cities."} IEEE Communications Magazine 52.8 (2014): 74-81.
\bibitem{unger16}		Unger, Moshe, et al.\textit{"Towards latent context-aware recommendation systems"},Knowledge-Based Systems, 2016, 104: 165-178.
\bibitem{Setten04}		Van Setten, Mark, Stanislav Pokraev, and Johan Koolwaaij. \textit{"Context-aware recommendations in the mobile tourist application COMPASS."} Adaptive hypermedia and adaptive web-based systems. Springer Berlin/Heidelberg, 2004.
\bibitem{Wang09}		Wang, Yi, et al. \textit{"framework of energy efficient mobile sensing for automatic user state recognition."} Proceedings of the 7th international conference on Mobile systems, applications, and services. ACM, 2009.
\bibitem{yuror11}		Yurur, O.; Moreno, Wilfrido.\textit{"Energy efficient sensor management strategies in mobile sensing"}, Proceedings of the ISTEC General Assembly, Porto Alegre, Brazil, 2011, 1620.
\bibitem{yuror13}		Yurur, Ozgur, et al.\textit{"Adaptive sampling and duty cycling for smartphone accelerometer"}, In: Mobile Ad-Hoc and Sensor Systems (MASS), 2013 IEEE 10th International Conference on. IEEE, 2013. p. 511-518.
\bibitem{zhang13}		Zhang, Lei, et al. \textit{"SensTrack: Energy-efficient location tracking with smartphone sensors"}, Addison Wesley, Massachusetts, IEEE sensors journal, 2013, 13.10: 3775-3784
\end{thebibliography}

\end{document}